\documentclass[conference]{IEEEtran}
\IEEEoverridecommandlockouts
\usepackage{cite}
\usepackage{enumitem}
\usepackage{amsmath,amssymb,amsfonts}
\usepackage{algorithmic}
\usepackage{graphicx}
\usepackage{textcomp}
\usepackage{xcolor}
\usepackage{float}
\usepackage{multirow}
\usepackage{booktabs}

\usepackage{adjustbox}
\usepackage{fontawesome}
\usepackage[most]{tcolorbox}
\usepackage{tabularx}
\usepackage{hyperref}
\usepackage{subcaption}
\usepackage[T2A, T1]{fontenc}
\usepackage[utf8]{inputenc}
\usepackage[english]{babel}

\usepackage{listings}
\usepackage{xcolor}

\lstdefinelanguage{json}{
    basicstyle=\ttfamily\footnotesize,
    commentstyle=\color{gray}\itshape,
    stringstyle=\color{blue},       
    columns=flexible,
    keepspaces=true,
    frame={},
    rulecolor=\color{black},
    backgroundcolor=\color{white},
    showstringspaces=false,
    alsoletter=_,
    breaklines=true,                
    breakatwhitespace=true,         
    postbreak=\mbox{\textcolor{red}{$\hookrightarrow$}\space}, 
    morekeywords={id,name,header,types,rows,question_id,table_id,question,sql,sel,conds,agg}
}

\lstdefinestyle{SQLStyle}{
    basicstyle=\ttfamily\footnotesize,
    commentstyle=\color{gray}\itshape, 
    showstringspaces=false,          
    columns=flexible,     
    keepspaces=true,
    frame={}, 
    rulecolor=\color{black},
    backgroundcolor=\color{white},
    breaklines=true,                
    breakatwhitespace=true,
    postbreak=\mbox{\textcolor{red}{$\hookrightarrow$}\space},
    morekeywords={SELECT, FROM, WHERE, AND, OR, GROUP BY, ORDER BY, AS, JOIN, ON, IN, NOT, NULL, IS, COUNT, SUM, AVG, MAX, MIN}
}

\lstdefinestyle{HighlightSQL}{
    basicstyle=\ttfamily\footnotesize,
    frame={},
    breaklines=true,
    breakatwhitespace=true,
    morestring=[b]",
    literate=
      {Attendance}{{\textbf{Attendance}}}1
      {Opponent}{{\textbf{Opponent}}}1
      {34348.0}{{\textbf{34348.0}}}1
      {38237.0}{{\textbf{38237.0}}}1
}

\def\BibTeX{{\rm B\kern-.05em{\sc i\kern-.025em b}\kern-.08em
    T\kern-.1667em\lower.7ex\hbox{E}\kern-.125emX}}
\begin{document}

\title{LLMSQL: Upgrading WikiSQL for the LLM Era of Text-to-SQL}

\author{
\IEEEauthorblockN{1\textsuperscript{st} Dzmitry Pihulski}
\IEEEauthorblockA{\textit{Department of Artificial Intelligence} \\
\textit{Wroclaw University of Science and Technology}\\
Wroclaw, Poland \\
dzmitry.pihulski@pwr.edu.pl}
\and
\IEEEauthorblockN{2\textsuperscript{nd} Karol Charchut}
\IEEEauthorblockA{\textit{Trusted Artificial Intelligence} \\
\textit{Wroclaw University of Science and Technology}\\
Wroclaw, Poland \\
karol.charchut.dev@gmail.com}
\and
\IEEEauthorblockN{3\textsuperscript{rd} Viktoria Novogrodskaia}
\IEEEauthorblockA{\textit{Trusted Artificial Intelligence} \\
\textit{Wroclaw University of Science and Technology}\\
Wroclaw, Poland \\
novogrodskaiaviktoria@gmail.com}
\and
\IEEEauthorblockN{4\textsuperscript{th} Jan Kocoń}
\IEEEauthorblockA{\textit{Department of Artificial Intelligence} \\
\textit{Wroclaw University of Science and Technology}\\
Wroclaw, Poland \\
jan.kocon@pwr.edu.pl} 
}

\maketitle

\begin{abstract}
Converting natural language questions into SQL queries (Text-to-SQL) enables non-expert users to interact with relational databases and has long been a central task for natural language interfaces to data. While the WikiSQL dataset played a key role in early NL2SQL research, its usage has declined due to structural and annotation issues, including case-sensitivity inconsistencies, data types mismatches, syntax errors, and unanswered questions. We present LLMSQL, a systematic revision and transformation of WikiSQL designed for the LLM era. We classify these errors and implement automated methods for cleaning and re-annotation. To assess the impact of these improvements, we evaluated multiple large language models (LLMs), including Gemma 3, LLaMA 3.2, Mistral 7B, gpt-oss 20B, Phi-3.5 Mini, Qwen 2.5, OpenAI o4-mini, DeepSeek-R1, and others. Notably, DeepSeek-R1 achieves 88.40\% accuracy in a zero-shot setting, and models under 10B parameters surpass 90\% accuracy after finetuning. Rather than serving as an update, LLMSQL is introduced as an LLM-ready benchmark: unlike the original WikiSQL—tailored for pointer-network models selecting tokens from input—LLMSQL provides clean natural-language questions and full SQL queries as plain text, enabling straightforward generation and evaluation for modern natural-language-to-SQL models.

\end{abstract}

\begin{IEEEkeywords}
Natural language processing (NLP), SQL query generation, language models (LLM), dataset quality assessment, cleaning and annotation, training on tabular data, WikiSQL, Text-to-SQL benchmarks
\end{IEEEkeywords}

\section{Introduction}
Converting natural language questions into structured queries is a fundamental task in developing intelligent data access systems. This capability enables users without technical expertise to retrieve information from databases using plain language, without the need to learn query languages such as SQL.

The WikiSQL dataset, comprising over 80{,}000 pairs of natural language questions and corresponding SQL queries with table data from Wikipedia, is one of the most widely used resources for training and evaluating text-to-SQL systems. It is widely recognized as a benchmark dataset in NL2SQL research. However, despite its popularity, WikiSQL suffers from several issues that limit its reliability in practical applications and rigorous research. These include data type mismatches, case sensitivity inconsistencies, and questions that do not return answers even when the generated queries are correct.

Beyond WikiSQL, subsequent benchmarks broadened the scope of Text-to-SQL evaluation. 
Spider~\cite{Spider2018} emphasizes cross-domain generalization with multi-table queries, while BIRD~\cite{Wretblad2024BIRD} focuses on robustness to annotation noise in real-world settings. 
These efforts illustrate complementary directions, yet systematic cleaning of WikiSQL itself has remained underexplored despite its historical importance and widespread use.

In most prior works, these issues have either been ignored or only partially addressed, leading to inflated performance metrics and poor generalization of models on real-world data. In this work, we propose a fundamentally different approach: a comprehensive reworking and cleaning of the WikiSQL dataset to produce a reliable and reproducible benchmark.

To guide the reader, the remainder of this paper is structured as follows. 
Section~\ref{sec:related_work} reviews related work, Section~\ref{sec:dataset} details our re-annotation of WikiSQL, and Section~\ref{sec:evaluation} presents evaluation with large language models. 
We conclude in Section~\ref{sec:conclusion}, with prompts and examples provided in the Appendix.

The main contributions of our work are as follows:

\begin{enumerate}
    \item We release the official evaluation package for the LLMSQL benchmark\footnote{\url{https://github.com/LLMSQL/llmsql-benchmark}}, which provides standardized scoring, execution-based evaluation, and reference utilities for researchers and practitioners.

    \item We conducted a systematic analysis of WikiSQL, identifying key issues affecting query execution accuracy, including type mismatches, case sensitivity inconsistencies, and unanswerable questions.
    
    \item We refined the dataset annotations through both manual and automated methods to ensure stable query execution and correct results.
    
    \item We introduce \textbf{LLMSQL}: an improved version of the WikiSQL dataset that preserves the original scale and diversity while significantly enhancing its practical usability for structured query generation tasks.
    
    \item We performed comparative experiments with multiple language models on both the original WikiSQL splits and the LLMSQL benchmark.
    
    \item We adapted LLMSQL to modern research needs, particularly for evaluation with large language models (LLMs). The dataset will be publicly released on major platforms following publication.
\end{enumerate}

In this landscape, LLMSQL is complementary to existing resources: 
it provides a large-scale, single-table benchmark with validated annotations, 
in contrast to Spider, which is best suited for evaluating multi-table and compositional reasoning, 
and BIRD, which targets robustness under noisy annotations. 
Researchers can thus choose the benchmark that best matches their evaluation goals.

Our work aims to enable more transparent, reliable, and practical research in natural language interfaces to databases, and provides a cleaned and well-validated resource for further model development and benchmarking.

\section{Related Work}
\label{sec:related_work}

Converting natural language questions into SQL (NL2SQL) is a central problem for natural database interfaces, with progress driven by large annotated corpora and neural architectures. A key dataset is \textbf{WikiSQL}, released with Seq2SQL (Zhong et al.~\cite{Zhong2017Seq2SQL}), which introduced execution-based reinforcement learning to optimize queries by both text and results. Seq2SQL’s pointer-network design—selecting tokens from the input—limits its applicability to modern LLMs.

Subsequent research focused on better capturing SQL structure and schema. SQLNet~\cite{Xu2017SQLNet} used a sketch-based, slot-filling decoder for SELECT, aggregation, and WHERE components. TypeSQL~\cite{Yu2018TypeSQL} added type-aware features linking question tokens to table columns, improving handling of rare values. Coarse-to-Fine methods first predict a skeleton, then detailed WHERE conditions. Execution-aware approaches like SeaD~\cite{Xu2021SeaD} and execution-guided decoding~\cite{Wang2018EG} prune invalid queries, improving execution accuracy, which remains the main metric for WikiSQL.

Execution-aware decoding techniques have also proven effective. Execution-guided decoding runs partial SQL against the target table during generation and prunes candidate continuations that lead to execution errors or empty/invalid results; this approach has been shown to reduce syntactic and semantic errors in generated queries and improve the share of successfully executable outputs~\cite{Wang2018EG}.

The integration of pretrained transformers with table-aware contextualization produced further gains beyond earlier seq2seq models. For instance, SQLova~\cite{Hwang2019SQLova} adapts BERT-style encoders to incorporate table headers and cell values, yielding stronger logical-form and execution accuracy. Complementary approaches enhance schema representations and alignment between natural language and table content. These advances illustrate that pretrained models can surpass execution-guided seq2seq baselines. However, such results hold only for single-table WikiSQL and should be interpreted with caution when generalizing to multi-table or cross-domain settings

Recognizing these limitations, new benchmarks were introduced. \textbf{Spider}~\cite{Spider2018} established a cross-domain standard with multi-table queries, JOINs, and nested structures, requiring compositional reasoning. \textbf{UNITE}~\cite{Lan2023UNITE} aggregated multiple datasets for broader SQL pattern coverage, while \textbf{BIRD}~\cite{Wretblad2024BIRD} emphasized robustness under noisy annotations. These resources expand scope, but none directly address the persistent quality issues in WikiSQL itself.

A critical body of work addresses annotation quality, evaluation methodology, and robustness. Finegan-Dollak et al.~\cite{FineganDollak2018} analyzed evaluation practices and demonstrated that certain dataset splits and anonymization choices can inflate reported performance; they proposed more conservative splitting strategies and evaluation protocols to better assess compositional generalization. Recent research, such as studies on BIRD-Bench, has highlighted that crowd-sourced datasets often contain issues such as noisy annotations, mismatched data types, and questions that cannot be answered from the table. These problems can bias standard evaluation metrics, including both execution accuracy and logical-form accuracy, giving a misleading impression of model performance~\cite{Wretblad2024BIRD}. To address this, researchers recommend cleaning datasets and performing explicit verification of model outputs. Additionally, broader benchmark analyses emphasize the need for standardized evaluation procedures and the inclusion of realistic, user-generated queries, which help ensure that models are tested on scenarios closer to real-world usage and that reported metrics more accurately reflect practical performance~\cite{EDBT2025Analysis,Lan2023UNITE}.

Taken together, the literature highlights three concurrent directions: (1) architectural and schema-aware improvements to increase generation correctness (e.g., type-aware models, table-aware encoding models, execution-guided decoding)~\cite{Hwang2019SQLova, Yu2018TypeSQL, Wang2018EG}; (2) the design of more representative and cleaner benchmarks and evaluation practices that mitigate annotation noise (e.g., Spider, UNITE, and the recommendations of Finegan-Dollak)~\cite{Spider2018, FineganDollak2018, Lan2023UNITE}; and (3) the systematic evaluation of large pretrained and/or instruction-tuned LMs for Text-to-SQL, along with the development of practical engineering techniques (prompting, constrained decoding, schema linking, execution validation) required to obtain executable and reliable queries in real-world settings~\cite{Liu2024SurveyLLM, Zhang2024BenchmarkLLM}.

In this work, we focus on directions (2) and (3): we perform a systematic analysis of WikiSQL annotation quality, apply both automatic and manual corrections (including normalization of case and spacing, data type validation, SQL syntax fixes, and removal of unanswerable examples), and evaluate the performance of modern LLMs on the resulting benchmark.

\section {Data}
\label{sec:dataset}

\subsection{Introduction to NL2SQL and the WikiSQL Dataset}

\textbf{WikiSQL} contains 80{,}654 (question, SQL) pairs derived from 24{,}241 Wikipedia tables~\cite{Zhong2017Seq2SQL}. It has become a standard benchmark due to its scale, simple query structures, and public availability. 

WikiSQL provides three file groups—tables, questions, and SQLite database files—organized into train, validation (dev / val), and test splits.

The dataset uses only two SQL data types, \texttt{real} and \texttt{text}—corresponding to \texttt{number} and \texttt{string} in JSON. Each table (Listing~\ref{lst:example_of_table}) contains several fields, including:

\begin{itemize}
    \item \texttt{id}: Unique table identifier.
    \item \texttt{header}: Column names.
    \item \texttt{types}: Data types (\texttt{real} or \texttt{text}).
    \item \texttt{rows}: Table entries.
    \item \texttt{name}: Optional table name derived from the ID.
\end{itemize}

\begin{lstlisting}[language=JSON,caption={Example Table from WikiSQL Dataset.},label={lst:example_of_table},captionpos=b]
{
  "id": "1-10088101-1",
  "name": "table_10088101_1",
  "header": [
    "No. in set","No. in series","Title","Directed by",
    "Written by","Original air date","Production code"
  ],
  "types": ["real","real","text","text","text","text","text"],
  "rows": [
    [1,174,"Per Manum","Kim Manners",
     "Chris Carter & Frank Spotnitz","February 18, 2001","8ABX13"],
    [2,175,"This is Not Happening","Kim Manners",
     "Chris Carter & Frank Spotnitz","February 25, 2001","8ABX14"],
    [9,184,"Nothing Important Happened Today II","Tony Wharmby",
     "Chris Carter & Frank Spotnitz","November 18, 2001","9ABX02"]
  ]
}
\end{lstlisting}

Questions in the JSONL files (Listing~\ref{lst:example_question}) include:

\begin{itemize}
    \item \texttt{table\_id}: Table referenced.
    \item \texttt{question}: Natural language query.
    \item \texttt{sql}: WikiSQL representation of the SQL query:
    \begin{itemize}
        \item \texttt{sel}: Index of the column selected.
        \item \texttt{conds}: \texttt{WHERE} conditions as [column index, operator (\texttt{=}, \texttt{>}, \texttt{<}), value].
        \item \texttt{agg}: Aggregation code: 0=none, 1=\texttt{MAX}, 2=\texttt{MIN}, 3=\texttt{COUNT}, 4=\texttt{SUM}, 5=\texttt{AVG}.
    \end{itemize}
\end{itemize}

\begin{lstlisting}[language=JSON,caption={Example Question from WikiSQL Dataset.},label={lst:example_question},captionpos=b]
{
  "table_id": "1-10088101-1",
  "question": "The episode with production code 9ABX02 was originally aired on what date?",
  "sql": {"sel":5,"conds":[[6,0,"9ABX02"]],"agg":0}
}
\end{lstlisting}

The SQL follows the general format (Listing~\ref{lst:SQL-format}), with the example query shown in Listing~\ref{lst:SQL-query-example}.

\begin{lstlisting}[style={SQLStyle},label={lst:SQL-format},caption={SQL Query Format.},captionpos=b]
SELECT agg_op agg_col
FROM table
WHERE cond1_col cond1_op cond1
    AND cond2_col cond2_op cond2 ...;
\end{lstlisting}

\begin{lstlisting}[style=SQLStyle,label={lst:SQL-query-example},caption={SQL Query for Question in Listing~\ref{lst:example_question}},captionpos=b]
SELECT "Original air date"
FROM "1-10088101-1"
WHERE "Production code" = '9ABX02';
\end{lstlisting}

WikiSQL excludes advanced SQL clauses such as \texttt{JOIN}, subqueries, or window functions.

\subsection{Identified Issues in the WikiSQL Dataset}

WikiSQL shows several structural and semantic issues that can affect model training and evaluation:

\begin{itemize}
    \item \textbf{Incomplete information}: Some tables ($\sim$140) miss one column name, which we augmented manually (common headers: ``Year'', ``Month'', ``Rank'', ``Position'').
    \item \textbf{Datatype conflicts}: Literals or row values sometimes mismatch their annotated type:
    \begin{itemize}
        \item Numbers stored as strings with commas or signs.
        \item \texttt{text} columns holding integers/floats.
        \item \texttt{real} columns holding integers.
    \end{itemize}
    \item \textbf{Duplicates}: Identical tables/questions with different IDs give false diversity.
    \item \textbf{Empty results}: 49.25\% of queries return no rows; 41.22\% due to case sensitivity. The remainder (8.03\%) likely involves other case issues.
    \item \textbf{Non-intuitive SQL format}: Numeric placeholders for column indices, aggregation, and operators obscure query structure and hinder human/LLM evaluation.
\end{itemize}

\subsection{Cleaning and Re-annotation Process}
\subsection*{Incomplete information}
Empty column names are not allowed in SQL tables. In the dataset, 140 tables were identified with a single missing column name each. To resolve this issue, the missing column names were manually annotated based on the stored values.
For instance, the missing column name in Listing~\ref{lst:empty_column_example} was replaced with \textit{"Prime Minister Ordinal Number(s)"}.
\begin{lstlisting}[language={JSON}, label={lst:empty_column_example}, caption={Part of Table with Missing Column Name.}, captionpos=b]
{
  "id": "1-11377572-3",
  "header": ["#", "", "Name", "Party", "Term in office", "The Times overall", "Matthew Parris", "Peter Riddell", "Ben MacIntyre"],
  "types": ["real", "text", "text", "text", "text", "real", "real", "real", "real"],
  "rows": [
    [1, "1", "Robert Walpole", "Whig", "1721–1742", 9, 14, 16, 7],
    [3, "3", "Henry Pelham", "Whig", "1743–1754", 29, 19, 34, 20],
    [42, "61 63", "Winston Churchill", "Conservative", "1940–1945 1951–1955", 1, 1, 1, 1],
    [50, "71", "Margaret Thatcher", "Conservative", "1979–1990", 5, 3, 4, 10],
    [52, "73", "Tony Blair", "Labour", "1997–2007", 16, 34, 15, 12]
  ]
}
\end{lstlisting}

\subsection*{Datatype Conflicts}
Most of the datatype issues were resolved automatically by addressing specific cases:

\begin{itemize}
    \item \textbf{\texttt{real} stored as strings}: Spaces, commas, and leading "+" or "-" signs were removed, and values were converted to \texttt{float} (\texttt{real}). Additionally, literals in questions were corrected to eliminate English numeral formatting. \\
    Example: "-2,123.9" => "-2139.9" => -2139.9.
    
    \item \textbf{\texttt{text} stored as float or int}: Values were written to \texttt{text} (string). \\
    Example: 1224 => "1224"
    
    \item \textbf{\texttt{real} stored as int}: Values were converted to \texttt{float} (\texttt{real}). \\
    Example: 628 => 628.0
\end{itemize}

Only a small number of cases required manual correction.

\subsection*{Duplicates}
Duplicates can negatively affect training, validation, and other downstream tasks when the dataset is used for such purposes. Therefore, they were systematically reduced as follows:
\begin{itemize}
    \item Duplicate tables were removed.
    \item Questions associated with removed tables were reassigned to the corresponding non-duplicated tables.
    \item Duplicate questions were removed.
    \item Tables with no associated questions were removed.
\end{itemize}

A table was considered a duplicate if it shared the same column names, column data types, and row values with another table. Similarly, a question was considered a duplicate if it had the same natural language query, table column names, and column data types as another question, since these are the inputs provided to the model during evaluation.

After cleaning, the resulting number of tables and questions is summarized in Table~\ref{tab:after_duplicates}.

\begin{table}[!h]
\centering
\caption{Number of Tables and Questions With and Without Duplicates}
\label{tab:after_duplicates}
\footnotesize
\begin{tabular}{lcc}
\toprule
\textbf{Duplicates} & \textbf{Tables} & \textbf{Questions} \\ 
\midrule
With       & 26,531 & 80,654 \\
Without    & 25,609 & 80,330 \\
\bottomrule
\end{tabular}
\end{table}

\subsection*{Empty Query Results}
The preprocessed data was used to retrieve results from the tables. 
An \emph{empty result} refers to queries that return no records: either an empty list (\texttt{[]} for plain \texttt{SELECT} queries without aggregation) or a placeholder \texttt{[[None]]} when aggregate functions such as \texttt{AVG}, \texttt{MIN}, \texttt{MAX}, or \texttt{SUM} are applied but no matching rows are found. 
Cases involving \texttt{COUNT} function, which returns \texttt{[[0]]} when no matches are present, were not included in this analysis.

Overall, 49.25\% of all queries produced such empty results. 
Of these, 41.22\% were conclusively caused by case mismatches in string literals — for these queries, adjusting the case ensured that results were returned. 
The case mismatch between literals in the \texttt{sql} key of a question's JSON and the corresponding literals in the \texttt{question} key occurs in the WikiSQL dataset. 
An example is shown in Listing~\ref{lst:case_mismatch_question}.

The case mismatch between literals in the \texttt{sql} key of a question's JSON and the corresponding literals in the \texttt{question} key occurs in the WikiSQL dataset. An example is shown in Listing~\ref{lst:case_mismatch_question}.

\begin{lstlisting}[language=JSON, caption={Example of Case Mismatch between SQL and Natural Language Query.}, label={lst:case_mismatch_question}, captionpos=b]
{
  "table_id": "1-10015132-11",
  "question": "What position does the player who played for butler cc (ks) play?",
  "sql": {
    "sel": 3,
    "conds": [
      [5, 0, "Butler CC (KS)"]
    ],
    "agg": 0
  }
}
\end{lstlisting}

\noindent
It is seen that the SQL literal is \texttt{Butler CC (KS)}, whereas in the question it appears as \texttt{butler cc (ks)}.  
For each question, SQL string literals were updated to match the case in the natural language query. However, this alone did not resolve most empty results.

If the result was still empty, all capitalization variants—changing the first letter of each word (where a word is defined as a substring separated by spaces) in the literals—were generated and tested until a non-empty result was returned. The algorithm proceeds as follows:
\begin{itemize}
    \item \textbf{Generate capitalization variants:} Create or variants as mentioned above. \\
Example for literals \texttt{['New York', 'Beijing']}:
\begin{lstlisting}[language={json}]
['New York', 'Beijing']
['new York', 'Beijing']
['new york', 'Beijing']
['new york', 'beijing']
['New york', 'Beijing']
['New york', 'beijing']
['New York', 'beijing']
\end{lstlisting}
    \item \textbf{Test SQL queries}: Create SQL queries using each combination of variants and execute until a non-empty result is found or all variants are tested.
    \item \textbf{Update literals}: If a non-empty result is found, update both the SQL and the natural language query literals accordingly.
\end{itemize}

For the remaining queries that returned no results, the case of string literals was adjusted to match the corresponding table values (applied only for the comparison operator '='). The procedure proceeds as follows:
\begin{itemize}
    \item \textbf{Locate literal in table}: For each SQL literal with "=" operator, identify the matching string in the table row.
    \item \textbf{Update case}: Adjust the literal's case to match the table value.
    \item \textbf{Test and update query}: Create a new SQL query using the updated literals. If the query returns a result, update the literals in both SQL and the natural language question.
\end{itemize}

The presented approach successfully eliminated empty results for the targeted portion of queries (41.22\%). The remaining 8.03\% were unaffected by case adjustments in the \texttt{questions} JSON files, and their underlying causes remain to be investigated in future work.

\subsection*{Non-intuitive SQL Format}
The original SQL storage (as shown in Listing~\ref{lst:example_question}) was replaced with standard SQL queries.  
By adhering to SQL standards and avoiding functions beyond regular clauses, the queries are now compatible with any SQL database following the standard.  
Additionally, numeric placeholders for column names and operators (both comparison and aggregation) were replaced with the actual names. The said mapping is displayed in Table~\ref{tab:mapping}.

\begin{table}[!h]
\centering
\caption{Mapping of Numeric Placeholders to Aggregation and Comparison Operators}
\label{tab:mapping}
\footnotesize
\begin{tabular}{c l l}
\toprule
\textbf{Placeholder} & \textbf{Aggregation Operator} & \textbf{Comparison Operator} \\ \midrule
0 & No operator & = \\
1 & MAX         & > \\
2 & MIN         & < \\
3 & COUNT       & — \\
4 & SUM         & — \\
5 & AVG         & — \\ 
\bottomrule
\end{tabular}
\end{table}

\noindent
Refined JSON does not include initial SQL description format---\texttt{sql} field includes already constructed query as shown in Listing~\ref{lst:refined_example_question}.

\begin{lstlisting}[language=JSON, caption={Refined Example Question.}, label={lst:refined_example_question}, captionpos=b]
{
  "question_id": 20,
  "table_id": "1-10088101-1",
  "question": "The episode with production code 9ABX02 was originally aired on what date?",
  "sql": "SELECT \"Original air date\" 
    FROM \"1-10088101-1\" 
    WHERE \"Production code\" = '9ABX02';"
}
\end{lstlisting}

\subsection{Additional Unresolved Issues}
Despite the refinements applied, several issues remain unresolved. A notable issue is the mismatch of aggregation operators. In certain cases, the annotated operator does not align with the intent of the natural language query or the contents of the table. Although the resulting SQL queries are syntactically valid, the choice of operator introduced by the annotator may produce semantically incorrect results.

The following example illustrates a case in which \texttt{COUNT} was used instead of \texttt{SUM}:

\textbf{Question:} \emph{How many people attended when the opponent was \textbf{twins}?}

\textbf{Original (Incorrect SQL):}
\begin{lstlisting}[style=SQLstyle]
SELECT COUNT("Attendance")
FROM "2-12206127-5"
WHERE "Opponent" = 'twins';
\end{lstlisting}

\textbf{Corrected SQL (aggregates \textbf{Attendance} properly):}
\begin{lstlisting}[style=SQLstyle]
SELECT SUM("Attendance")
FROM "2-12206127-5"
WHERE "Opponent" = 'twins';
\end{lstlisting}

\textbf{Schema and Sample Rows (key column: \textbf{Attendance}):}
\begin{itemize}[leftmargin=*]
    \item \textbf{Types:} ["text", "text", "text", "text", \textbf{"real"}, "text"]
    \item \textbf{Columns:} ["Date", "Opponent", "Score", "Loss", \textbf{"Attenfdance"}, "Record"]
    \item \textbf{Sample Rows:} \textbf{Attendance} values are 34348.0, 38237.0, and 23849.0; values of matching rows should be summed to obtain the final result.
\end{itemize}

\begin{lstlisting}[
  frame={}, 
  basicstyle=\ttfamily\footnotesize, 
  breaklines=true, 
  breakatwhitespace=true,
  postbreak=\mbox{\textcolor{red}{$\hookrightarrow$}\space}
]
["July 1", "Red Sox", "4 - 0", "Carpenter (7-5)", 34348.0, "38-43"]
["July 2", "Red Sox", "16 - 4", "Loaiza (5-9)",   38237.0, "38-44"]
["July 31", "Twins", "3 - 1", "Mays (12-8)", 23849.0, "49-58"]
\end{lstlisting}

\textbf{Query Results:}
\begin{itemize}
    \item \texttt{COUNT}: \texttt{[[0]]}
    \item \texttt{SUM}: \texttt{[[None]]}
\end{itemize}

Although \texttt{SUM} is intended to aggregate attendance values, the query returned \texttt{[[None]]}. This occurs because, according to the previously defined criteria for empty results, \texttt{[[0]]} for \texttt{COUNT} was not considered empty. If the string literal had matched the case in the table (e.g., "Twins" instead of "twins"), the correct result would have been 23849.0.

Assigning the correct aggregation operator is challenging to assess automatically and is most reliable when evaluated individually by considering the semantic meaning of the natural language query. Full automation is difficult due to the need for contextual understanding.

It remains under discussion whether all queries with no matching rows should be corrected, or if some should be left unmodified to retain diversity in the dataset. 
Table~\ref{tab:aggregators_empty} provides a summary of queries with non-matching rows, broken down by aggregation operator.

\begin{table}[!h]
\centering
\caption{Distribution of Empty Results by Aggregation Operator}
\label{tab:aggregators_empty}
\footnotesize
\begin{tabular}{lrr}
\toprule
\textbf{Aggregation} & \textbf{Nominal} & \textbf{Percentage} \\ 
\midrule
No operator          & 290   & 0.36\% \\
\texttt{MAX}         & 1,569 & 1.95\% \\
\texttt{MIN}         & 1,562 & 1.94\% \\
\texttt{COUNT}       & 1,577 & 1.96\% \\
\texttt{SUM}         & 1,461 & 1.82\% \\
\texttt{AVG}         & 1,571 & 1.96\% \\
\textbf{TOTAL}       & \textbf{8,030} & \textbf{10.00\%} \\
\bottomrule
\end{tabular}
\end{table}


\section{LLM Evaluation on Different Scenarios}
\label{sec:evaluation}

The primary motivation for the modifications to the WikiSQL dataset described in Section~\ref{sec:dataset} was to create a modern, ready-to-use benchmark for large language models. Unlike the original WikiSQL, LLMSQL enables evaluation of LLMs that generate queries based on knowledge acquired during pre-training and/or instruction-tuning. Such LLMs do not require additional fine-tuning to produce well-formatted, executable SQL queries that satisfy the input question.

Providing LLMSQL as a ready-to-use benchmark, we revive a classic resource that had largely been abandoned. Along with the dataset, we propose a set of standardized rules for evaluation, ensuring that models assessed under these rules can be fairly compared. A key element in this process is prompt design. We created our own prompt and corresponding few-shot examples for LLM evaluation across different scenarios. For each few-shot example, we additionally included a sample row from an imaginary table crafted to match the example, and for the actual question we also added the first row of the real target table. We found that models can greatly benefit from such additions, as this helps them better understand the structure and content of the table. The complete prompt and five-shot examples are provided in Appendix~\ref{prompt_shots}.

For instance, a 1-shot prompt for a simple count query might look like:

\begin{tcolorbox}[colback=white,colframe=black!70, title=LLMSQL 1-shot Prompt, boxrule=0.5pt, breakable]
\textbf{Instruction:} You are an expert SQLite SQL query generator. Your task: given a question and a table schema, output \textbf{ONLY} a valid SQL \texttt{SELECT} query.

\textbf{Strict Rules:}
\begin{itemize}
    \item Output \textbf{only} SQL (no explanations, no markdown, no \texttt{```} fences)
    \item Use table name \texttt{"Table"}
    \item Allowed functions: \texttt{MAX, MIN, COUNT, SUM, AVG}
    \item Allowed condition operators: \texttt{=, >, <, !=}
    \item Allowed SQL keywords: \texttt{SELECT, WHERE, AND}
    \item Always use quotes for all column names and table name, even single-word names (e.g., \texttt{"Price"}, \texttt{"Something \#"})
\end{itemize}
\textbf{Example:}\newline
Question: What is the price of the Samsung Galaxy S23?\newline
Columns: ['Brand', 'Model', 'Price', 'Storage', 'Color']\newline
Types: ['text', 'text', 'real', 'text', 'text']\newline
Sample row: ['Apple', 'iPhone 14', 899.99, '128GB', 'White']\newline
SQL: SELECT "Price" FROM "Table" WHERE "Brand" = "Samsung" AND "Model" = "Galaxy S23";
\newline\newline
Question: \{question\} \newline
Columns: \{headers\}\newline
Types: \{types\}\newline
Sample row: \{sample\_row\}\newline
SQL:

\end{tcolorbox}

\subsubsection{Temperature}

For most models, the primary evaluation parameter was the temperature, which was set to zero (greedy decoding) to ensure reproducibility of results. 

Recent studies have shown that using greedy decoding to evaluate reasoning models can lead to higher repetition rates and significant variability across different checkpoints~\cite{deepseek_r1}. Therefore, for models with strong reasoning capabilities, we did not use greedy decoding. Instead, we employed the default sampling parameters recommended for each model.

\subsubsection{Maximum Number of Tokens}

For all models, the maximum number of newly generated tokens was limited to 256. This limit was chosen arbitrarily, as all SQL queries in our benchmark typically require no more than approximately 50 tokens. The additional margin ensures that models producing intermediate reasoning steps or verbose outputs can still generate the final query without being truncated.

\subsubsection{Evaluation Metric}
\label{eval_metric}
The evaluation metric used in this study is \textit{execution accuracy}. A generated query is considered correct if, when executed against the database, it produces the same result set as the ground truth query. Specifically, the result must contain the same number of rows, identical columns, and matching values (regardless of the order of results).

It is also important to specify the database system used for query execution. We employed \texttt{SQLite}, accessed via the standard Python client in version 3.11.11. The choice of SQLite was motivated by two factors: (i) the original WikiSQL dataset was designed for this system, and (ii) SQLite is included in the Python standard library, ensuring easy reproducibility. 

However, this choice introduces some limitations, as different SQL dialects provide varying sets of functions and syntax. In the case of LLMSQL, these limitations are minimal because the benchmark does not require complex functions. All queries are formulated using basic SQL constructs that are supported by virtually all SQL dialects, which makes the benchmark portable to other database systems if needed.

\subsubsection{SQL Extraction from Outputs}

Not all models strictly adhere to the prompt specification given in Appendix~\ref{prompt_shots}. Many models exhibit a tendency to first restate the task or produce intermediate steps before generating the final SQL query. In such cases, the SQL query might not appear at the beginning of the response, but rather in the middle or at the end.  

To address this, we adopted a regex-based extraction strategy similar to what is used in popular text-to-SQL benchmarks such as Spider~\cite{Spider2018}. Specifically, we designed a regex pattern to extract all possible SQL statements from the model output:

\begin{verbatim}
(?s)SELECT\b.*?(?=(?:;|\n{4,}|```|$))
\end{verbatim}

\textbf{Regex explanation:}
\begin{itemize}
    \item \texttt{(?s)} -- Enables single-line (DOTALL) mode, so the dot (\texttt{.}) matches newline characters as well.
    \item \texttt{SELECT\b} -- Matches the keyword \texttt{SELECT} followed by a word boundary.
    \item \texttt{.*?} -- Lazily matches any characters (including newlines) until the next condition.
    \item \texttt{(?=(?:;|\textbackslash n\{4,\}|```|\$))} -- Positive lookahead that ensures the match stops at one of the following:
        \begin{itemize}
            \item A semicolon (\texttt{;})
            \item Four or more consecutive newlines (\texttt{\textbackslash n\{4,\}})
            \item A Markdown code block fence (\texttt{```})
            \item The end of the string (\texttt{\$})
        \end{itemize}
\end{itemize}

For instance, a model might output:

\begin{tcolorbox}[colback=white,colframe=black!70, title=Example Regex Extraction, boxrule=0.5pt, breakable]
\textbf{Output:}
First, we need to filter employees:
SELECT Name FROM employees WHERE Salary > 50000;
Then count them...

\textbf{Extracted:}
SELECT Name FROM employees WHERE Salary > 50000;

\end{tcolorbox}

After extracting all substrings that match the regex, we limit the number of candidate SQL queries to the first 10 occurrences. This constraint prevents excessive generation and mitigates potential gaming strategies by models, such as producing numerous variants to increase the chance of a correct answer. The value of 10 was chosen arbitrarily to balance completeness and computational efficiency.

Subsequently, all $n$ queries ($n \leq 10$) were executed against the database. If at least one query produced the same results as the ground truth (as defined by the evaluation metric in Section~\ref{eval_metric}), the model's response was considered correct.

\subsection{Main Scenario (Benchmark)}
\label{main_bench}

\begin{table*}[hbt!]
\centering
\caption{Performance of LLMs on the LLMSQL benchmark under 0-shot, 1-shot, and 5-shot settings. Models are ordered by parameter size.}
\label{benchmark_results}
\footnotesize
\begin{tabular}{lrrrrrr}
\toprule
\textbf{Model} & \textbf{Parameters} & \textbf{0-shot (1000)} & \textbf{1-shot (1000)} & \textbf{5-shot} & \textbf{Eval. Time (min)} & \textbf{Temperature} \\ 
\midrule
Llama 3.2 1B Instruct & 1.2B & 5.70\% & 4.00\% & 22.44\% & 20  & 0.0 \\
Qwen 2.5 1.5B Instruct & 1.5B & 20.60\% & 53.70\% & 53.41\% & 20  & 0.0 \\
Phi 3.5 mini Instruct & 3.8B & 24.70\% & 57.20\% & 62.07\% & 40  & 0.0 \\
Gemma 3 4B IT & 4.3B & 60.90\% & 59.10\% & 64.29\% & 40  & 0.0 \\
Mistral 7B Instruct v0.2 & 7.2B & 24.40\% & 50.70\% & 56.99\% & 40  & 0.0 \\
gpt-oss-20b & 21.5B & 47.30\% & 67.10\% & 71.77\% & 50  & 1.0 \\
DeepSeek V3 0324 & 685B & 77.30\% & 79.90\% & 83.52\% & API & API default \\
DeepSeek R1 0528 & 685B & \textbf{88.40\%} & \textbf{88.30\%} & \textbf{86.57\%} & API & API default \\
OpenAI o4-mini & Proprietary & 85.50\% & 85.60\% & 86.45\% & API & API default \\
\bottomrule
\end{tabular}
\end{table*}

Since LLMSQL is based on the relatively simple WikiSQL dataset, our main evaluation method does not rely on the original dataset splits. Instead, the evaluation uses all questions from 3 subsets, which were not filtered out during dataset cleaning. We encourage the use of such a strategy in the future, since it will bring more robust results of evaluation.

The evaluation results for the LLMSQL benchmark are presented in Table~\ref{benchmark_results}. The benchmark includes three settings: \textbf{0-shot}, \textbf{1-shot}, and \textbf{5-shot}. We used 1000 randomly chosen set of examples to evaluate the models on 0-shot and 1-shot settings in favor of saving resources. As the 1-shot prompt, we used the first example from the 5-shot prompt in the Appendix~\ref{prompt_shots}.

We evaluated a broad set of models across different scenarios, including \texttt{Llama 3.2 1B Instruct}, \texttt{Qwen 2.5 1.5B Instruct}, \texttt{Phi 3.5 Mini Instruct}, \texttt{Gemma 3 4B IT}, \texttt{Gemma 3 27B IT}, \texttt{Mistral 7B Instruct v0.2}, \texttt{gpt-oss-20B}, \texttt{DeepSeek V3 0324}, \texttt{DeepSeek R1 0528}, and \texttt{OpenAI o4-mini}~\cite{llama3_paper, qwen2.5, phi3.5, gemma_2025, mistral_7b, gpt_oss, deepseek_v3, deepseek_r1, openai_o4_mini}.

Model performance generally improves with size, but not strictly monotonically. For example, Gemma 3 (4.3B) achieves 60.9\% in 0-shot, outperforming Mistral 7B, which scores only 24.4\%, indicating that architectural differences and instruction tuning play a major role beyond parameter count.

Across all models, accuracy consistently improves when moving from 0-shot to 1-shot and further to 5-shot settings. For example, Phi 3.5 (3.8B) increases from 24.7\% in the 0-shot setting to 62.07\% in the 5-shot setting, representing a relative improvement of approximately 151\%. Similarly, gpt-oss-20B improves from 47.3\% to 71.77\%. These results suggest that providing examples helps models better adapt to the specifics of the dataset. This is particularly important in our scenario because models tend to \emph{overshoot} the complexity of the task. 

Upon analyzing model outputs, we observed that many models generated unnecessarily complex SQL queries, often including subqueries and aliases—structures that are not present in LLMSQL. Additionally, some models frequently used unsupported functions such as \texttt{SUBSTRING()}, which caused execution failures since SQLite does not include this function in its standard set.

Among the smallest models in our benchmark, Llama 3.2 (1.2B) and Qwen 2.5 (1.5B), we observe a significant performance gap despite their relatively close parameter counts. Llama 3.2 achieves only 5.7\% accuracy in the 0-shot setting and improves to 22.44\% with 5-shot, indicating that it struggles severely with SQL generation even when provided with examples. In contrast, Qwen 2.5 starts at 20.6\% in 0-shot and reaches 53.41\% in 5-shot, more than doubling its accuracy compared to Llama under identical conditions. This discrepancy suggests that model architecture and training data quality are critical, as parameter size alone does not explain the difference.

The largest proprietary and open source models dominate the LLMSQL benchmark. In particular, DeepSeek R1 0528 achieves the highest scores across all settings, reaching up to 88.4\%, while OpenAI o4-mini closely follows with 86.45\% in 5-shot. These models outperform smaller open-source models by a substantial margin, often 30–60 percentage points. Interestingly, the largest models exhibit a performance plateau: for example, DeepSeek R1 slightly decreases from 88.4\% in 0-shot to 86.57\% in 5-shot, indicating that additional in-context examples provide limited improvement. Since the evaluation prompt explicitly specifies which SQL functions are allowed and which SQL dialect to use, even in the 0-shot setting, this plateau suggests that larger models are able to quickly adapt to task-specific instructions and generate correct SQL queries without relying heavily on few-shot examples.

\subsection{Additional Scenario: Fine-Tuning}

\begin{figure}
  \includegraphics[width=\columnwidth]{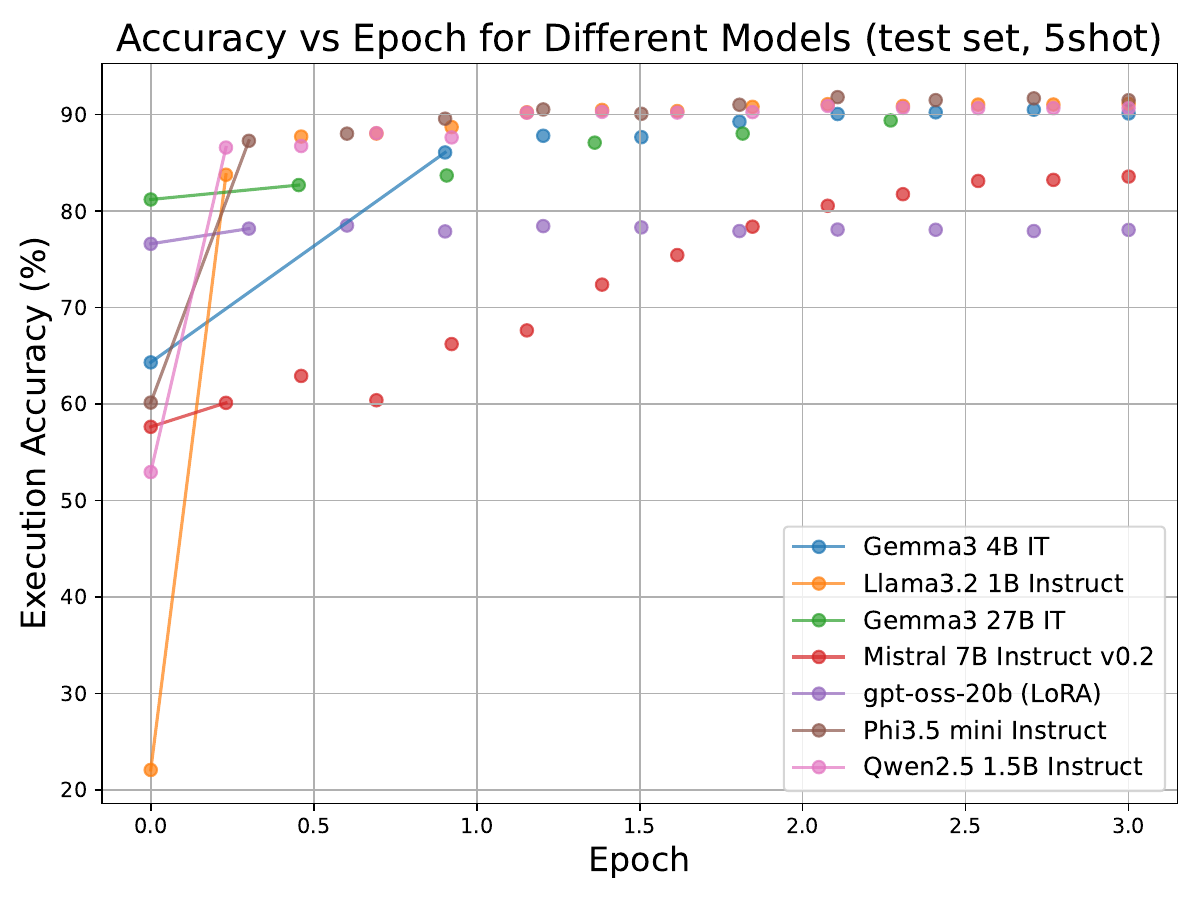}
  \caption{Execution accuracy on the LLMSQL test split during fine-tuning for various models in the 5-shot setup. The x-axis represents fine-tuning epochs, and the y-axis shows execution accuracy (\%).}
  \label{fig:finetune_test_results}
\end{figure}

The main scenario (Section~\ref{main_bench}) demonstrated that many models struggle to fully understand the task from the prompt alone. In particular, they often fail to correctly handle the set of allowed SQL functions and operators without extensive few-shot prompting.

To provide a broader perspective on model capabilities, we also evaluated a fine-tuning scenario using the LLMSQL benchmark. For this, LLMSQL was divided into three splits—train, validation, and test—following the original WikiSQL split assignments as closely as possible. Each question remains in the same split as in the original dataset.

In this scenario, models are fine-tuned on the training split using Cross-Entropy Loss and evaluated dynamically on the test split according to the execution accuracy metric described in Section~\ref{eval_metric}. This allows us to observe how performance evolves with the number of fine-tuning steps.

All models were trained under consistent conditions, with fixed numbers of epochs, learning rates, and other hyperparameters. The only variable across models was batch size, which was adjusted according to model capacity. Fine-tuning hyperparameters for the experiments were as follows:

\begin{itemize}
    \item \textbf{Batch size:} 16--128 (depending on model capacity)
    \item \textbf{Optimizer:} \texttt{adamw\_bnb\_8bit}
    \item \textbf{Learning rate:} $4\mathrm{e}{-5}$
    \item \textbf{LR scheduler:} cosine
    \item \textbf{LoRA (for gpt-oss-20B):} rank $r=8$, $\alpha=16$
    \item \textbf{LoRA target modules:} all linear layers, specifically
    \texttt{7.mlp.experts.gate\_up\_proj}, \texttt{7.mlp.experts.down\_proj}, 
    \texttt{15.mlp.experts.gate\_up\_proj}, \texttt{15.mlp.experts.down\_proj}, 
    \texttt{23.mlp.experts.gate\_up\_proj}, \texttt{23.mlp.experts.down\_proj}
\end{itemize}

The resulting execution accuracy on the test split as a function of fine-tuning steps is shown in Figure~\ref{fig:finetune_test_results}. The fine-tuning process was stable for all models, with no observed loss spikes or gradient norm anomalies.

We also included the Gemma3 27B IT model to provide a comparison with a model of similar parameter size to gpt-oss-20B.

The fine-tuning results varied across models. Smaller models—including Gemma3 4B IT, Llama3.2 1B Instruct, Phi3.5 Mini Instruct, and Qwen2.5 1.5B Instruct—benefited greatly from fine-tuning, achieving more than 90\% execution accuracy on the test split. This indicates that fine-tuning substantially improved their understanding of dataset-specific structures and conventions, making them more flexible and effective for NL2SQL tasks.

For larger models (Gemma3 27B IT, Mistral 7B Instruct v0.2, and gpt-oss-20B LoRA), improvements were observed but were generally smaller; no model achieved over 90\% execution accuracy. Notably, gpt-oss-20B had relatively stable performance around 78--78.5\% execution accuracy despite LoRA fine-tuning, suggesting that larger models may require more specialized tuning strategies or longer fine-tuning schedules to fully leverage dataset-specific structures.

We note that the chosen fixed training conditions may not have been optimal for all models. For instance, the execution accuracy did not begin to decrease for any model, suggesting that three epochs may not have been sufficient for the models to reach their maximal performance. The purpose of using fixed conditions was to place all models on an equal footing, ensuring that the evaluation focused on their ability to learn from the dataset itself rather than on extensive hyperparameter tuning. All specific hyperparameters used for fine-tuning and evaluation, along with each model's fine-tuning curves, will be released upon revision of the paper.

\section{Conclusions}
\label{sec:conclusion}

Our motivation was to address longstanding issues in the original dataset. By applying both automated normalization and targeted manual corrections, we created a resource that is structurally sound, semantically consistent, and practically executable across all examples. LLMSQL stores complete and human-readable SQL statements, making it directly suitable for modern language models without additional preprocessing.

Interestingly, while larger reasoning-oriented LLMs such as DeepSeek R1 and OpenAI o4-mini achieved execution accuracies above 85\%, our finetuning experiments revealed that even relatively small models can surpass 90\% accuracy when adapted to LLMSQL.

We also argue that, despite its simplicity, LLMSQL remains a relevant and informative benchmark. Real-world SQL workloads are often dominated by relatively simple patterns: in an analysis of 8.1 million production queries at Uber~\cite{johnson2018practical}, over 62\% used \texttt{JOIN}, and fewer than 1\% involved operators such as \texttt{UNION}, \texttt{INTERSECT}, or \texttt{EXCEPT}. This reinforces the idea that mastering simple clauses is critical for practical deployment and benchmarking progress.

We hope LLMSQL will revitalize interest in lightweight, high-quality datasets and serve as a reliable resource for advancing natural language interfaces to databases, both as a standalone benchmark and as part of broader evaluation suites.

\section{Future Work}
Although LLMSQL in its current version already solves the key problems of the original dataset, further development of the resource is important both for improving quality and expanding its scope of application. Future steps will aim to make the data set more complete, complex, and diverse while maintaining its structural simplicity and practical applicability. In particular, the following areas of work are planned:

\begin{itemize}
    \item \textbf{Adding questions to tables.}  
    In the current version, some tables do not have related questions. Creating additional annotations will provide more complete data coverage and increase the value of the dataset as a training resource.

    \item \textbf{Introduction of JOIN queries.}  
    To increase the complexity and practical usefulness of the dataset, it is planned to add SQL queries that include table joins. This step will bring the tasks closer to real-world business applications, where JOIN operations are standard practice.

    \item \textbf{Addition of new data types.} 
    Future versions will include additional information formats, such as dates and time values. This will allow models to handle a wider range of conditions (e.g., date range filters, time interval calculations), bringing tasks closer to real-world SQL usage.

    \item \textbf{Multilingual support.}  
    Translating questions and table descriptions into several languages will expand the user audience and allow exploration of model behavior in a multilingual SQL generation environment.

    \item \textbf{Scaling and integration with other benchmarks.}  
    Moreover, several existing benchmarks are not fully suited for modern LLMs, as they often lack clearly defined input–output pairs or require complex preprocessing. We plan to adapt them and combine with LLMSQL. This will provide greater diversity of structures and scenarios, allow models to work with a wider range of queries, and maintain the coherence and consistency of annotations.
\end{itemize}

Together, these enhancements should transform LLMSQL into a resource that combines compactness and ease of use with high informational value and practical utility. We expect that the updated dataset will be in demand both in academic research and in applied projects related to building natural language interfaces to databases.

\section*{Acknowledgments}
This work was financed by
(1) the European Regional Development Fund as part of the 2021–2027 European Funds for a Modern Economy (FENG) programme, project no. FENG.02.04-IP.040004/24: CLARIN – Common Language Resources and Technology Infrastructure;
(2) the European Regional Development Fund as a part of the 2014-2020 Smart Growth Operational Programme, project no. POIR.04.02.00-00C002/19: CLARIN – Common Language Resources and Technology Infrastructure;
(3) Digital Research Infrastructure for the Arts and Humanities DARIAH-PL (POIR.04.02.00-00-D006/20, KPOD.01.18-IW.03-0013/23);
(4) CLARIN ERIC – European Research Infrastructure Consortium: Common Language Resources and Technology Infrastructure (period: 2024-2026) funded by the Polish Minister of Science under the programme: "Support for the participation of Polish scientific teams in international research infrastructure projects", agreement number 2024/WK/01;
(5) Polish Minister of Digital Affairs under a special purpose subsidy No. 1/WII/DBI/2025: HIVE AI: Development and Pilot Deployment of Large Language Models in the Polish Public Administration;
(6) the statutory funds of the Department of Artificial Intelligence, Wroclaw University of Science and Technology.

AI-based tools, including ChatGPT, Grammarly Premium, and Writeful, were used exclusively to support linguistic clarity and improve the readability of the manuscript.

\bibliographystyle{IEEEtran}
\bibliography{IEEEfull, conference_101719}

\appendix

\section{Prompt and Few-Shot Examples}
\label{prompt_shots}

We provide the prompt and five-shot examples used for evaluating LLMs on LLMSQL. The LLM is instructed to generate valid SQL queries given a question and table schema, following strict rules.
\begin{tcolorbox}[colback=white,colframe=black!70, title=LLMSQL Prompt and Examples, boxrule=0.5pt, breakable]
\textbf{Instruction:} You are an expert SQLite SQL query generator. Your task: given a question and a table schema, output \textbf{ONLY} a valid SQL \texttt{SELECT} query.

\textbf{Strict Rules:}
\begin{itemize}
    \item Output \textbf{only} SQL (no explanations, no markdown, no \texttt{```} fences)
    \item Use table name \texttt{"Table"}
    \item Allowed functions: \texttt{MAX, MIN, COUNT, SUM, AVG}
    \item Allowed condition operators: \texttt{=, >, <, !=}
    \item Allowed SQL keywords: \texttt{SELECT, WHERE, AND}
    \item Always use quotes for all column names and table name, even single-word names (e.g., \texttt{"Price"}, \texttt{"Something \#"})
\end{itemize}

\textbf{Few-Shot Examples:}

\textbf{Example 1:}
Question: What is the price of the Samsung Galaxy S23?
Columns: ['Brand', 'Model', 'Price', 'Storage', 'Color']
Types: ['text', 'text', 'real', 'text', 'text']
Sample row: ['Apple', 'iPhone 14', 899.99, '128GB', 'White']
SQL: SELECT "Price" FROM "Table" WHERE "Brand" = "Samsung" AND "Model" = "Galaxy S23";

\textbf{Example 2:}
Question: How many books did Maya Chen publish?
Columns: ['Author', 'Books Published', 'Genre', 'Country', 'Years Active']
Types: ['text', 'real', 'text', 'text', 'text']
Sample row: ['John Smith', 3, 'Non-fiction', 'Canada', '2005–2015']
SQL: SELECT "Books Published" FROM "Table" WHERE "Author" = "Maya Chen";

\textbf{Example 3:}
Question: What is the total population of cities in California?
Columns: ['City', 'State', 'Population', 'Area', 'Founded']
Types: ['text', 'text', 'real', 'real', 'text']
Sample row: ['Houston', 'Texas', 2304580, 1651.1, '1837']
SQL: SELECT SUM("Population") FROM "Table" WHERE "State" = "California";

\textbf{Example 4:}
Question: How many restaurants serve Italian cuisine?
Columns: ['Restaurant', 'Cuisine', 'Rating', 'City', 'Price Range']
Types: ['text', 'text', 'real', 'text', 'text']
Sample row: ['Golden Dragon', 'Chinese', 4.2, 'Boston', '\$\$']
SQL: SELECT COUNT(*) FROM "Table" WHERE "Cuisine" = "Italian";

\textbf{Example 5:}
Question: What is the average salary for Software Engineers?
Columns: ['Job Title', 'Salary', 'Experience', 'Location', 'Company Size']
Types: ['text', 'real', 'text', 'text', 'text']
Sample row: ['Data Analyst', 70000, 'Junior', 'Chicago', '200–500']
SQL: SELECT AVG("Salary") FROM "Table" WHERE "Job Title" = "Software Engineer";

Question: \{question\} 
Columns: \{headers\}
Types: \{types\}
Sample row: \{sample\_row\}
SQL:

\end{tcolorbox}

\end{document}